\documentclass[preprint,12pt]{elsarticle}

\usepackage{url}
\usepackage{xcolor}
\usepackage{color}

\usepackage{amssymb}
\usepackage{amsmath}
\usepackage{ulem}
\usepackage{multirow}
 
\biboptions{comma,square,numbers,sort&compress}

\newcommand{\cb}[1]{{\color{black} #1}}

\journal{Transportation Research Part C}

\begin{document}

\begin{frontmatter}

%% Title, authors and addresses

%% use the tnoteref command within \title for footnotes;
%% use the tnotetext command for the associated footnote;
%% use the fnref command within \author or \address for footnotes;
%% use the fntext command for the associated footnote;
%% use the corref command within \author for corresponding author footnotes;
%% use the cortext command for the associated footnote;
%% use the ead command for the email address,
%% and the form \ead[url] for the home page:
%%
%% \title{Title\tnoteref{label1}}
%% \tnotetext[label1]{}
%% \author{Name\corref{cor1}\fnref{label2}}
%% \ead{email address}
%% \ead[url]{home page}
%% \fntext[label2]{}
%% \cortext[cor1]{}
%% \address{Address\fnref{label3}}
%% \fntext[label3]{}

\title{Explainable, automated urban interventions to improve pedestrian and vehicle safety}

%% use optional labels to link authors explicitly to addresses:
%% \author[label1,label2]{<author name>}
%% \address[label1]{<address>}
%% \address[label2]{<address>}

\author[uoc]{C. Bustos\corref{cor1}}
\ead{mbustosro@uoc.edu}

\author[uoc]{D. Rhoads\corref{cor1}}
\ead{drhoads@uoc.edu}

\author[uoc]{A. Sol\'e-Ribalta}

\author[uoc]{D. Masip}

\author[urv]{A. Arenas}

\author[uoc,mit]{A. Lapedriza}

\author[uoc]{J. Borge-Holthoefer\corref{cor1}}
\ead{jborgeh@uoc.edu}

\cortext[cor1]{Corresponding author}
\address[uoc]{Internet Interdisciplinary Institute (IN3), Universitat Oberta de Catalunya, \\Barcelona 08860, Catalonia, Spain}
\address[mit]{Media Lab, Massachusetts Institute of Technology, 02139 Cambridge, MA}
\address[urv]{Departament d'Enginyeria Inform\`atica i Matem\`atiques, Universitat Rovira i Virgili, \\43007 Tarragona, Spain}

\begin{abstract}
%% Text of abstract
\cb{At the moment, urban mobility research and governmental initiatives are mostly focused on motor-related issues, e.g. the problems of congestion and pollution. And yet, we can not disregard the most vulnerable elements in the urban landscape: pedestrians, exposed to higher risks than other road users. Indeed, safe, accessible, and sustainable transport systems in cities are a core target of the UN's 2030 Agenda. Thus, there is an opportunity to apply advanced computational tools to the problem of traffic safety, in regards especially to pedestrians, who have been often overlooked in the past. This paper combines public data sources, large-scale street imagery and computer vision techniques to approach pedestrian and vehicle safety with an automated, relatively simple, and universally-applicable data-processing scheme. The steps involved in this pipeline include the adaptation and training of a Residual Convolutional Neural Network to determine a hazard index for each given urban scene, as well as an interpretability analysis based on image segmentation and class activation mapping on those same images. Combined, the outcome of this computational approach is a fine-grained map of hazard levels across a city, and an heuristic to identify interventions that might simultaneously improve pedestrian and vehicle safety. The proposed framework should be taken as a complement to the work of urban planners and public authorities.}
\end{abstract}

%The quest for safe, affordable, accessible and sustainable transport systems in cities is a core target of the UN's 2030 Agenda for Sustainable Development. While concerns about congestion and pollution capture a lot of attention, special care is needed for the most vulnerable elements in the urban landscape: pedestrians.

\begin{keyword}
Deep Learning \sep Google Street View \sep Mapillary \sep Pedestrian \sep Traffic safety
%% keywords here, in the form: keyword \sep keyword

%% MSC codes here, in the form: \MSC code \sep code
%% or \MSC[2008] code \sep code (2000 is the default)

\end{keyword}

\end{frontmatter}

%%
%% Start line numbering here if you want
%%
%\linenumbers

% Guide for authors:
% https://www.elsevier.com/journals/transportation-research-part-c-emerging-technologies/0968-090x/guide-for-authors (see Preparation section)

%% main text
\section{Introduction}
\label{sec:intro}
%State the objectives of the work and provide an adequate background, avoiding a detailed literature survey or a summary of the results.
In the last century, the accelerated growth of urban areas has given rise to challenges at a variety of levels. Among these, mobility stands out. The ability to efficiently move people and goods is critical to a city's social and economic success \cite{de2014navigability,jiang2016timegeo,abbar2018structural}. It is unsurprising, then, the enormous amount of economic and engineering effort that urban planners have devoted to enhance the efficiency of road networks, bus lines, and metro systems \cite{GAKENHEIMER1999671}. Unlike transportation modes that operate in exclusive spaces, such as metro lines, the uncontrolled rise in urban automotive mobility has gone hand in hand with the degradation of other modes of transportation. Of all these alternative modes, walking has suffered the most, due in large part to the fact that the amount of the streetscape allotted to vehicles invades and interferes with the pedestrian space. Nevertheless, cities exhibit a growing tendency to stop and reverse this process by fostering more active, citizen-friendly transportation modes --foot, bike and personal mobility vehicles, which compete for this public space  \cite{cervero2003}.   

%While the positive outcomes of this new mobility paradigm away from driving are quantifiable, e.g. implications for citizen health and pollution reduction [], it also comes with a sensitive side-effect: a parallel rise in pedestrian injuries and fatalities [].
%While we do not intend to analyze here the environmental \cite{sole2018decongestion}, public health \cite{nieuwenhuijsen2016car}, demographic \cite{newman2011peak}, or economic \cite{klein2017millennials,abbar2018structural} causes of this paradigm shift, 

One logical consequence of this paradigm shift, is the increased level of interaction between pedestrians and motor vehicles, largely due to the overlapping use of common (or adjacent) spaces such as roads, sidewalks, and zebra-crossings. Such increase gives rise to an important, negative side-effect: a growth in pedestrian injuries and fatalities. Data from the National Highway Traffic Safety Administration (NHTSA) of the United States indicate that the number of pedestrian fatalities per year is rising in the U.S. \cite{fars2019}. After a steady decline from the mid-1990's to a low in 2009, there has been a clear and consistent reversal until 2017 (the last year of available data), when pedestrian fatalities surpassed a previous 23-year high in 1995. 

Traditionally, pedestrian safety research has focused on the impact of structural factors (e.g. road lanes \cite{ukkusuri2012role}, traffic network structure \cite{rifaat2011effect,moeinaddini2014relationship}, existence of direct line-of-sight between objects \cite{mecredy2012neighbourhood,fu2019investigating}, etc.). In addition, socio-behavioral factors may be concomitant, e.g. the change of individual behavior related to the use of new, distraction-causing technologies \cite{nasar2008mobile}, inside and outside of vehicles, which is not likely to diminish in the future. Also, demographic variables (socio-economic status, race, gender) may play a role as well \cite{mukoko2019examining}. Nonetheless, crashes that involve motor vehicles and pedestrians are understudied, and, at the micro level, much less so outside intersections \cite{hu2018dangerous}. 

\begin{figure}[h!]
  \begin{center}
        \includegraphics[width=0.7\columnwidth]{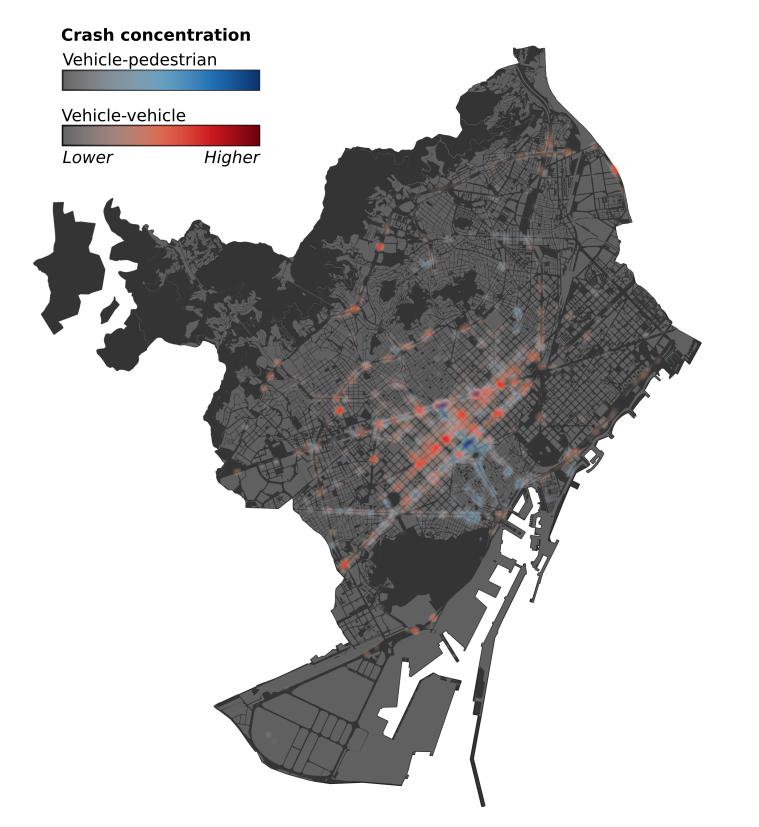}
  \end{center}
  \caption{{\bf Accident distribution in Barcelona.} Relative concentration of accidents by type (vehicle-to-pedestrian, vehicle-to-vehicle).}
  \label{fig1}
\end{figure}

%\begin{figure}[h!]
%  \begin{center}
%		\includegraphics[width=0.9\columnwidth]{figures/w-accidents-ped.png}
%	\end{center}
%	\caption{{\bf Accident distribution along street sections in Barcelona.} This figure illustrates the average proportion of total vehicle-pedestrian accidents occurring at different points (intersection, middle, etc.) along street segments. Each line represents an increasingly narrow filter of street lengths (from all streets to 200 meters +). For longer streets, there is a notable increase in the number of accidents in the middle of the street segment (a W formation in the redder lines). Intuitively, this might indicate a critical street length at which pedestrians are willing to risk injury to cross the street at unmarked points. The same plots (and behavior) can be seen in Figure~S3 of the SI.\rev{same plots for what? other cities?}}
%	\label{fig1}
%\end{figure}

An enlightening example, built upon real accident data, is shown in Figure~\ref{fig1}. Quite clear even to the naked eye, accidents involving vehicles may happen throughout a city. However, when a distinction is introduced (vehicle-to-vehicle {\it vs.} vehicle-to-pedestrian), the spatial patterns where these accidents occur are mostly non-overlapping, suggesting that the configuration of the public space --the scene where the accident happens-- matters, see as well Figure~S1 in the Supplementary Information (SI). All in all, the strategies for the safe coexistence of pedestrians and vehicles demand a separate and careful examination.

The combination of increasingly available street-level imagery sources and city open data portals, together with advances in the field of computer vision and larger training datasets \cite{zhou2014learning,zhou2017places}, has opened up promising new opportunities for facing challenges in urban science. Examples include the quantification of physical change and pattern identification in cities \cite{naik2017pnas,albert2017using,seiferling2017green}, \cb{road safety assessment \cite{song2018farsa}}, the prediction of human-perceived features of street scenes \cite{naik2014streetscore,liu2017machine}, the automated estimation of demographic variables across the United States \cite{gebru2017pnas} and Great Britain \cite{suel2019measuring}, or the beautification of urban images through the generation of prototypes \cite{kauer2018mapping}. Turning to transportation research, however, computer vision has focused mostly on traffic control and surveillance \cite{fadlullah2017state}, and automatic detection and collision prevention \cite{zhang2016faster,zhang2016far} for autonomous vehicles. Outside scene analysis, the Deep Learning paradigm has been exploited mostly on motor traffic \cite{polson2017deep,wu2018hybrid,zhang2018deep,wang2019enhancing,zhang2019multistep} , so far leaving aside its potential to tackle pedestrian safety.

Here, we address the complexities of vehicle-to-pedestrian interaction combining the structural (scene elements) and perceptual (scene composition) aspects of the problem. \cb{Overall, the contributions of the present work can be summarized as follows:
\begin{enumerate}
\item Creating a dataset of urban street-level images labelled according to accidentality, based on open data municipal accident records.
\item Developing a deep learning architecture, adapted from Deep Residual Networks (ResNet), for hazard index estimation in urban images, that works for both pedestrian and vehicle accidents, and is capable of producing city-wide hazard level landscapes at an unprecedented resolution of one value every 15-20 meters.
\item Proposing a set of interpretability analyses to extract human meaning from the outputs of the classification, through customized implementations of Pyramid Scene Parsing networks (PSPNet), Gradient-weighted class activation mapping (GradCam++), radar plots, and a new measure of scene disorder. 
\item Designing a greedy heuristic to propose realistic urban interventions, based on scene segmentation, class activation mapping and k-nn algorithm, which constitutes an informed guide for planners to pedestrian safety improvements.
\end{enumerate}
Taken together, these points constitute a novel and comprehensive deep learning pipeline for estimating vehicle and pedestrian hazard in urban scenes, and recommending feasible physical improvements to make those same scenes safer. The building blocks of the pipeline are tailored variants of different state-of-the-art deep learning/machine learning models and techniques (Deep Residual Networks (ResNet), Pyramid Scene Parsing network (PSPNet), Gradient-weighted class activation mapping (GradCam++)).

The remainder of the paper is organized as follows: in Section 2, data (collection, processing techniques and labelling) and methods (pipeline components) are described in detail; then, in Section 3, the results on the hazard index and landscape, its connection to scene composition, and intervention heuristic are presented and discussed. Finally, Section 4 summarizes the work and discusses possible gaps and lines of development.}

%%%%%%%%%%%%%%%%%%%%%%%%%%%%%%%%%%%%%%%%%%%%%%%%%%%%%%%%%
%%%%%%%%%%%%%%%%%%%%%%%%%%%%%%%%%%%%%%%%%%%%%%%%%%%%%%%%%
\section{Materials and Methods}
%Provide sufficient details to allow the work to be reproduced by an independent researcher. Methods that are already published should be summarized, and indicated by a reference. If quoting directly from a previously published method, use quotation marks and also cite the source. Any modifications to existing methods should also be described.
In this Section we provide the details about the datasets and Deep Learning methods that are used throughout the work. For an introduction to the Deep Learning paradigm, with a focus on transportation systems, we refer to Wang {\it et. al.} \cite{wang2019enhancing}. 
\subsection{Dataset collection and curation} 

To feed the proposed framework, we use two types of real urban data: historical accident statistics and street-level urban imagery.

In the case of Madrid and Barcelona, historical accident records for the years 2010-2018 are available from the open data portals of the respective municipal governments \cite{mad2019acc,bcn2019acc}. For San Francisco, data was available from 2015-2017 and it was filtered from the University of California, Berkeley's Transport Injury Mapping System (TIMS) of California traffic accidents \cite{tims2019}.
In total, the Barcelona dataset was made up of 86,414 accidents, 10,240 being pedestrian and 76,174 being vehicle accidents. The Madrid dataset had 76,026 accidents (12,533 pedestrian, 63,492 vehicle). In San Francisco, the dataset was made up of 15,492 accidents (3331 pedestrian, 12,161 vehicle). All data points are geolocated with their corresponding GPS coordinates. Besides location, due the detonating causes may be different, we distinguish between accidents where a vehicle and a pedestrian were involved (simply `pedestrian', or $P$, onwards), from vehicle-to-vehicle accidents (simply `vehicle', or $V$, onwards). The spatial distribution of empirical accident data for both vehicles and pedestrians can be seen in the SI Figure~S1.

Street-level imagery was extracted from two data sources. The Google StreetView (GSV) \cite{anguelov2010google} API was used for Barcelona and Madrid. In these dataset, images are, on average, 15 meters away from each other. As we wanted to capture the view of the driver, we limited our queries to images facing directly down the direction of traffic of the street. The result of this process was a comprehensive and homogeneous set of images for both cities. 

For the city of San Francisco, images were provided by Mapillary \cite{mapillary2019}, a crowd-sourced alternative to GSV. With Mapillary, all user-uploaded images are available under the CC-BY-SA license. As images are uploaded by private individuals working with different equipment, different setup, different light conditions, different vehicles, and without central coordination, several distinct challenges were presented by this dataset. Firstly, for each point provided, usually a single image was available. Occasionally, this image did not fit our criteria of facing down the direction of traffic, and had to be discarded. Secondly, data was only available from a smaller part of the city, corresponding to the area covered by the Mapillary contributors. The part of San Francisco available in the dataset, consisting mostly of high-traffic streets, is shown in Figure~S2 of the SI.

Combining data from different sources (GSV and Mapillary) allows us to test the robustness of our methods when dealing with similar, but not equally distributed, data . All the collected images, both for GSV and Mapillary, contain GPS locations in their metadata, which allows us to assign each street image a binary accident category (``safe'' vs. ``dangerous'').  We categorize a point as ``dangerous'' if one or more accidents have occurred with a 50 meter radius of its location. Otherwise, the point is categorized as ``safe''.  \cb{More details on the creation of the image dataset can be found in Section S1 of the SI, along with a more extended discussion of the trade-offs of using a radius to assign accidents to images in Section S4.}

The large collection of images tagged according to accident category was divided in 6 different datasets, resulting from the combination of the three targeted cities and two accident types ($V$ and $P$). The characteristics of each dataset (number of images per dataset and category) are detailed in Table~\ref{tab:datasets_img}. 

Notice that the San Francisco datasets are much smaller than Barcelona and Madrid datasets. For the 6 datasets, data was randomly split into train and test sets, containing $90\%$ and $10\%$ of the images respectively. 

\begin{table*}[th]
  \centering
\begin{tabular}{l|c|c|c|c|c}
\hline
{} & {} & \multicolumn{2}{c}{{\bf Vehicle} ($V$)} & \multicolumn{2}{c}{{\bf Pedestrian} ($P$)}\\
\hline
\hline
City & Total & Accident  & No accident & Accident  & No accident \\
\hline
\hline
Barcelona & 177645 & 61.8\% & 38.2\% &  48.1\% & 51.9\% \\
Madrid & 704950 & 48.3\%  & 51.7\% & 29.1\% & 70.9\% \\
San Francisco & 162530 & 35.7\% & 64.3\% & 17.4\% & 82.6\% \\
\hline
\end{tabular}
  \caption{Image dataset properties. Comparing the relative proportion of points with and without accidents across the various cities. In all 3 cities, there is a higher proportion of points with vehicle-to-vehicle accidents than vehicle-to-pedestrian accidents. Relatively less accident points in San Francisco reflects the smaller amount of accident data for that city.} %While Madrid and Barcelona had similarly sized accident datasets, there are significantly more street-level images and points in Madrid, leading to a balance between accident and no accident points. \asr{NO S'ENTEN. REVISAR VOLUMS DEL DATASET I COM S'HA BALANCEJAT EL CONJUNT}}
  \label{tab:datasets_img}
\end{table*}

\subsection{Hazard index estimation with Deep Learning} 

\cb{
A variety of Deep Learning architectures have shown to be remarkably effective for many computer vision tasks \cite{lecun2015deep,schmidhuber2015deep}. In this work we use a Residual Neural Network (ResNet) \cite{resnet_v2}, a particular architecture of Convolutional Neural Network (CNN), to estimate the {\it hazard index} ($H$) in new, unseen images. The main characteristic of ResNets is the implementation of ``shortcut connections'' that skip blocks of convolutional layers, allowing the network to learn residual mappings between layers that mitigate the vanishing gradients problem. For this critical step, all of the elements used were created from scratch -- training and test datasets, weight learning stage, etc. -- as is detailed in the following.

We define our {\it hazard index} ($H$) as the probability that a target image is classified as `dangerous' by the ResNet. For this objective, we train the ResNet to first classify images between the two defined accident categories: `dangerous' and `safe'. For each street-level image, the classifier delivers a value $H$ in the range of $[0,1]$. When $H \approx 1$, the point where the image was taken is considered as dangerous. On the contrary, when $H \approx 0$, the corresponding point is considered as safe. The hazard index is defined as the output of the Softmax activation function (between 0 and 1) of the last layer of the classifier architecture: 

\begin{equation}
H = \frac{e^{z_i}}{\sum_{j=0}^{K}e^{z_j}}  
\label{hazard}
\end{equation}
where $z$ is the output logits of the last ResNet layer, $i$ is the index of `dangerous'  class and $K$ is the number of classes. $H$ can be interpreted as the probability that the point related to a given image is hazardous.

To successfully train our ResNet architecture for the required classification task, we start with a pre-trained network that considers the Imagenet dataset~\cite{imagenet}, and then, via 'Transfer learning' techniques, we fine-tune the network using our data. At this stage, we remove the connections from the last layer of the pre-trained ResNet model, replace it with a new layer with two outputs (categories \textsl{dangerous} and \textsl{safe}), and randomly initialize the layer's weights. We re-trained (fine-tuned) this last layer, leaving the rest of the CNN static. To compensate for class imbalance during training stage, class weights were adjusted in the objective cross entropy loss function according to inverse class frequency:

\begin{equation}
w_i = \frac{1}{ln(c+r_i)} \\
\label{weights}
\end{equation}
with $w_i$ as the weight assigned to each class, $c$ is a parameter to control the range of the valid values, and $r_i$ is the ratio of the number of samples from each class respect the total of samples, and then

\begin{eqnarray}
Loss = \frac{1}{N}  \sum_{i=1}^{N}
w_i \cdot (y_i \cdot log(\hat{y_i})+(1-y_i) \cdot log(1-\hat{y_i}))
\label{loss}
\end{eqnarray}
where $N$ is the number of samples, and $y_i$ and $\hat{y_i}$ are the true label and the prediction for $i$ class, respectively.
In accordance with the defined accident types ($V$ and $P$), we train our ResNet to estimate two subtypes of hazard index: $H_V$ and $H_P$, corresponding to the hazard indices for vehicle-to-vehicle and vehicle-to-pedestrian accidents, respectively. Therefore, we end up training 6 models in total, two per city.}
%As examples, see the scores in the central columns of Figure~\ref{fig:CamAndSegmentation}. 

%\subsection{Image segmentation and Class Activation Maps} 
\subsection{Hazard index interpretability} 
One of the main shortcomings of Deep Learning techniques is (the lack of) interpretability. Certainly, deep neural networks can provide a high level of discriminative power, but at the cost of introducing many model variables, which eventually hinders the interpretability of their black-box representations \cite{adadi2018peeking}. \cb{This difficulty is especially pertinent in our case: improving pedestrian safety sometimes demands changes in the urban landscape, the question being {\it which} changes are pertinent. Here, we address this by using two different interpretability techniques. The first, scene disorder, is used to assess image complexity and the second, Class Activation Mapping (CAM), to assess which areas are more informative for the estimation of the hazard index. In particular, CAM methods have been recently shown to be successful for interpretability tasks in several fields \cite{fukui2019attention,wagner2019interpretable,desai2020ablation,patro2019u}, including medicine \cite{wang2017diabetic}.}

\subsubsection{Urban scene segmentation and scene disorder} 
First, in order to identify what objects are in the scene, and where they are positioned, we use urban scene segmentation. The goal of the semantic image segmentation task is to assign a category label to each pixel of an image. Segmentation provides a comprehensive breakdown of the physical elements visible in the scene. It predicts the label, location and mask for each object.
For this task, we used a high-performance method called Pyramid Scene Parsing Network (PSPNet) \cite{pspnet} architecture, pre-trained with the Cityscapes dataset  \cite{cityscapes}. PSPNet is a state-of-the-art deep learning model that exploits the capability of both global and local context information aggregation through several pyramid pooling layers. It has shown outstanding performance on several semantic segmentation benchmarks. Cityscapes is a real-world, vehicle-egocentric dataset for semantic urban scene understanding which contains 25K pixel-annotated images taken in different weather conditions. Images in Cityscapes are annotated with 30 urban object categories, but we used a subset of those (19) in our image repository segmentation --those that are common and relevant in driver-perspective scenes (e.g. ``car'', ``road'', ``sidewalk'', ``person'', ``traffic light'', etc.; see right-most labels in Figure~\ref{fig:CamAndSegmentation}).
  
  %The dataset defines 19 categories containing typical objects belonging to urban scenes such as car, road, sidewalk, person, traffic light, vegetation, among others, Also, these categories can be clustered in category groups. All the categories and groups are show in Table~\ref{tab:cityscapes}. PSPNet trained on the Cityscapes dataset allows us to detect, localise and segment 19 types of objects with an accuracy larger than $80\%$.

On top of the image segmentation outcome, we propose a measure of scene disorder inspired by the gray-tone spatial-dependence matrix \cite{haralick1973textural}, also known as Gray-level co-occurrence matrix (GLCM), which captures the amount of transitions between adjacent pixels labelled with different categories. It is known that complex images (related to scene disorder) may cause a division of attention \cite{moray1959attention,kahneman1973attention,alvarez2004capacity,richards2010development} and, as a consequence, reduce attention towards objects that are relevant to urban hazard.

\begin{figure}[h]
  \begin{center}
		\includegraphics[width=\columnwidth]{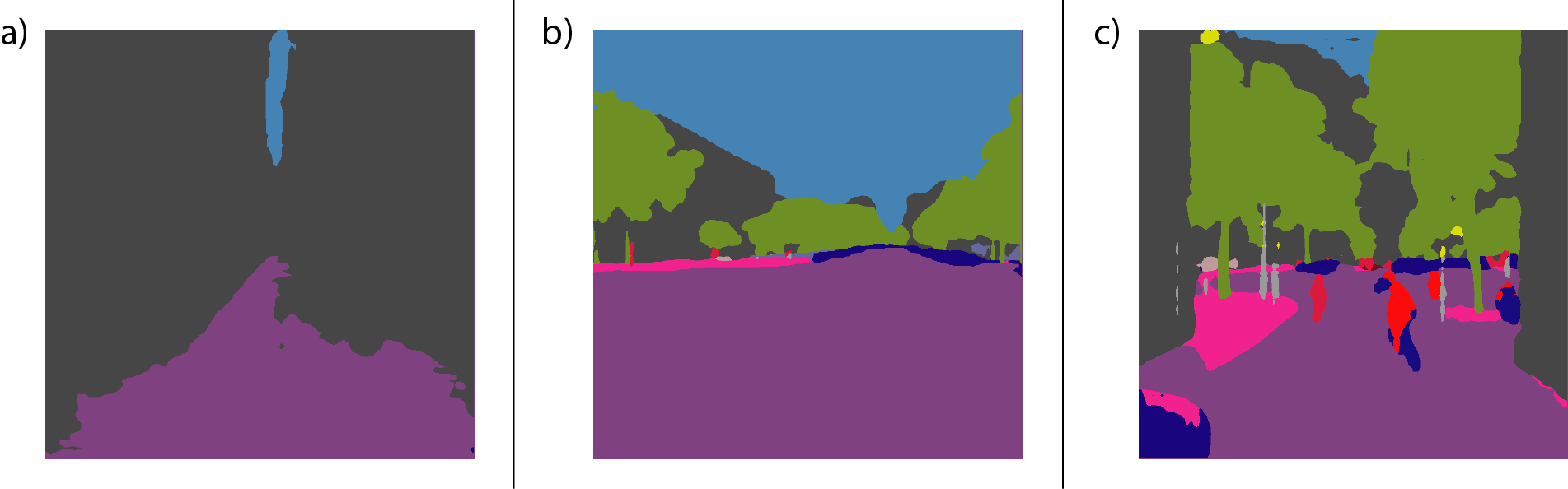}
	\end{center}
	\caption{{\bf Illustrating the concept of scene disorder.} Segmented images with low $SD = 0.15$ scene disorder (a); mild $SD = 0.39$ scene disorder (b); and high scene disorder $SD = 0.81$ (c).}
	\label{fig:spatial_disorder}
\end{figure}

Originally, GLCM characterizes the texture of an image by calculating how often pairs of pixels with specific values are adjacent in a specified spatial configuration. In our measure of scene disorder, the frequency of pair of pixels of different values is calculated over the segmented image, where the value of a pixel corresponds to an urban object category, instead of a gray intensity like the usual GLCM. We perform the calculation as follows:

\begin{equation}
SD =  \sum_{i=0}^{m}\sum_{j=0}^{n} 
\delta \left[ I(i,j) \neq I(i + \Delta i , j + \Delta j) \right]
\label{spadisorder}
\end{equation}
where $\delta[x]$ is the Kronecker delta, valued 1 if the condition $x$ is met, and 0 otherwise; and $\Delta i$ and $\Delta j$ represent an offset of 1, to compute the amount of pixel value transitions in two directions (right and below). With this definition, the measure $SD$ is incremented by 1 for every pair of neighboring pixels that have differing values. Examples of scene disorder measures can be seen in Figure~\ref{fig:spatial_disorder}.

 \subsubsection{Interpretability through Activation Mapping}
Moving on to the second step of our interpretability process, Class Activation Mapping (CAM) \cite{cam} and related techniques (e.g.~gradient-weighted class activation mapping (GradCAM++) \cite{gradcam,gradcamplus}) are used to interpret, visually, the patterns of images that are informative of a specific image category \cite{ventura2017interpreting,adadi2018peeking}, meaning, in our case, the regions that have influenced the most about the decision taken by the classifier for a certain class, in our case, classifying an image as 'dangerous'.
 
GradCAM++ was used to identify the regions of the image that are dangerous. Given an input image and a our trained CNN model, GradCAM++ generates a localization map by the use of the gradient information of the specific target class 'dangerous' to compute the target class weights of each feature map of the last convolutional layer of the CNN before the final classification. The final localization map is synthesized from the aggregated sum of these target class weights. Generating a GradCAM++ map for the 'dangerous' class helps to visually identify the specific patterns and objects learned by the CNN in order to differentiate between 'safe' and 'dangerous' scenes.  Since the images have been fully segmented, we can retrieve the objects that overlap with the dangerous regions. Analyzing frequencies, we can recover what object categories are more relevant to determine $H_V$ or $H_P$. Figure~\ref{fig:CamAndSegmentation} shows one example per city in the first column and visualizations of the described techniques in the other columns. In particular, second and third column display $H_P$ and $H_V$, respectively, with the corresponding Class Activation Map. Areas in red color are those that are more relevant to the hazard index, that is, areas that strongly contribute to increase the hazard indexes. Last column shows the automatic segmentation of the images.

%The main contribution of GradCAM++ compared to similar methods is the introduction of pixel-wise weighting of the gradients of the output regards to a particular spatial position in the final convolutional feature map of the CNN. This approach provides a measure of importance of each pixel in a feature map towards the overall decision of the CNN.

\subsection{A greedy heuristic to improve $H$}
\label{sec:greedy}
\cb{The combination of the Class Activation Mapping and image segmentation described in the previous section gives us insight into which regions and objects of a scene contribute most to its estimated hazard level. While this information is already relevant, it provides users with no concrete recommendations for structural changes to the scene that might make it safer. Accordingly, as a final step in the pipeline, we propose a strategy to exploit the large pool of images available in order to identify, for each scene, realistic and potentially low-cost physical alterations that would diminish $H_P$ and $H_{V}$ the most.}

\begin{figure*}[h!]
  \begin{center}
		\includegraphics[width=1\columnwidth]{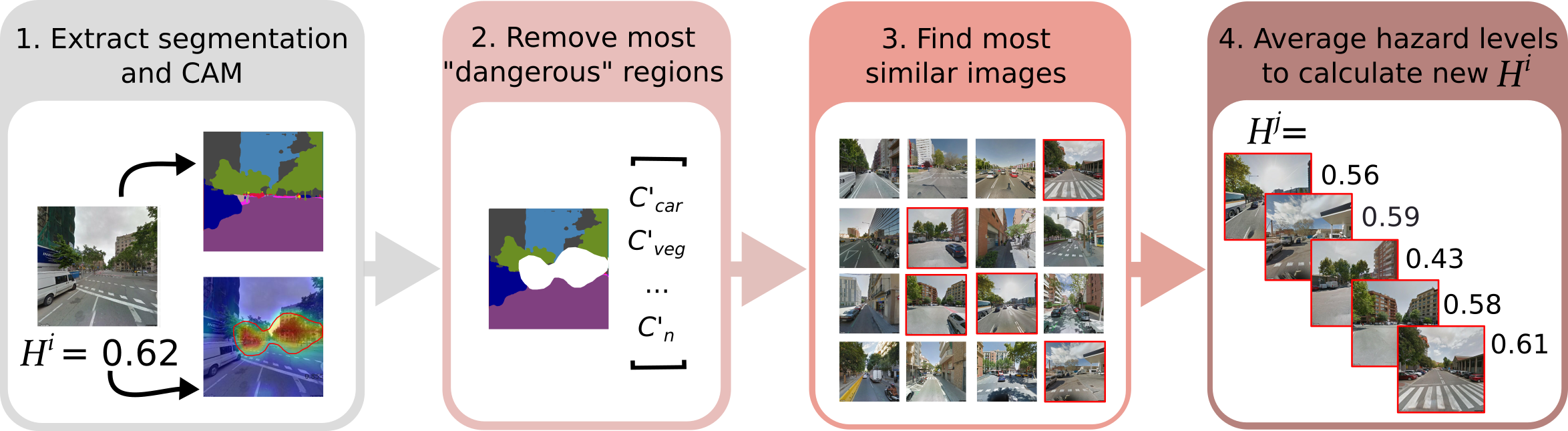}
	\end{center}
	\caption{{\bf Image hazard reduction flowchart.} Processing pipeline to improve the most hazardous parts of a street-level image $i$, comparing the new image with similar partner images $j$, and arriving at a new $H_P$ and $H_{V}$ for the original image.}
	\label{fig:process}
\end{figure*}

To this end, we take advantage of the methodologies developed in the previous steps. On the one hand, the segmentation task allows us to identify which objects among $C$ categories are present in a given scene (and to what extent). On the other, CAM provides information regarding which regions of the scene contribute most to the estimated hazard score. With this information at hand, for every image $i$ we build a vector of characteristics $v_i \in \mathbb{R}^{C}$, containing information of the relative area of category $C$ in $i$. For the target scene (the one for which we intend to reduce the hazard levels), we construct an additional surrogate vector of characteristics, $\tilde{v}_i$, in which we discard those regions that contribute most to $H_P$, i.e.~we only consider regions of $i$ where the class activation is mild-to-low ($< 0.7$), see first and second blocks in Figure~\ref{fig:process}. Next, we deploy an exhaustive search to find the five mirror images $j$ for $\tilde{v}_i$, with their respective vectors of characteristics $v_j$, such that their hazard index is lower:

\begin{eqnarray}
	& \mbox{argmin}_{j} ||\tilde{v}_i - v_j||_{2}\\
	& H^{j}_P < H^{i}_P \nonumber \\
	& H^{j}_V < H^{i}_V \nonumber
	\label{eq_mirror}
\end{eqnarray}
In other words, we seek the most similar locations in the city that have smaller $H_P$ and $H_{V}$ than $i$, see Fig.~\ref{fig:process} for a schematic representation of the process. The search for mirror images is limited to structurally similar scenes (compared to the original one), in order to promote simple and feasible interventions. \cb{We emphasize that this strategy is designed to be used in tandem with human users, who will be able to judge which recommendations are realistic. The choice of five images allows for some diversity in the range of interventions recommended.}

Finally, we remark that our approach is very similar to the regressive $k$-nearest neighbor ($k$-nn) algorithm \cite{harrington2012machine}, as opposed to a more sophisticated, Deep Learning-based mechanism for image ``safe-fication'' (following the concept of ``beautification'' in ref. \cite{kauer2018mapping}). These techniques lie beyond the scope of the present work.

\section{Experiments and Results}

\begin{figure*}[h!]
  \begin{center}
		\includegraphics[width=1\columnwidth]{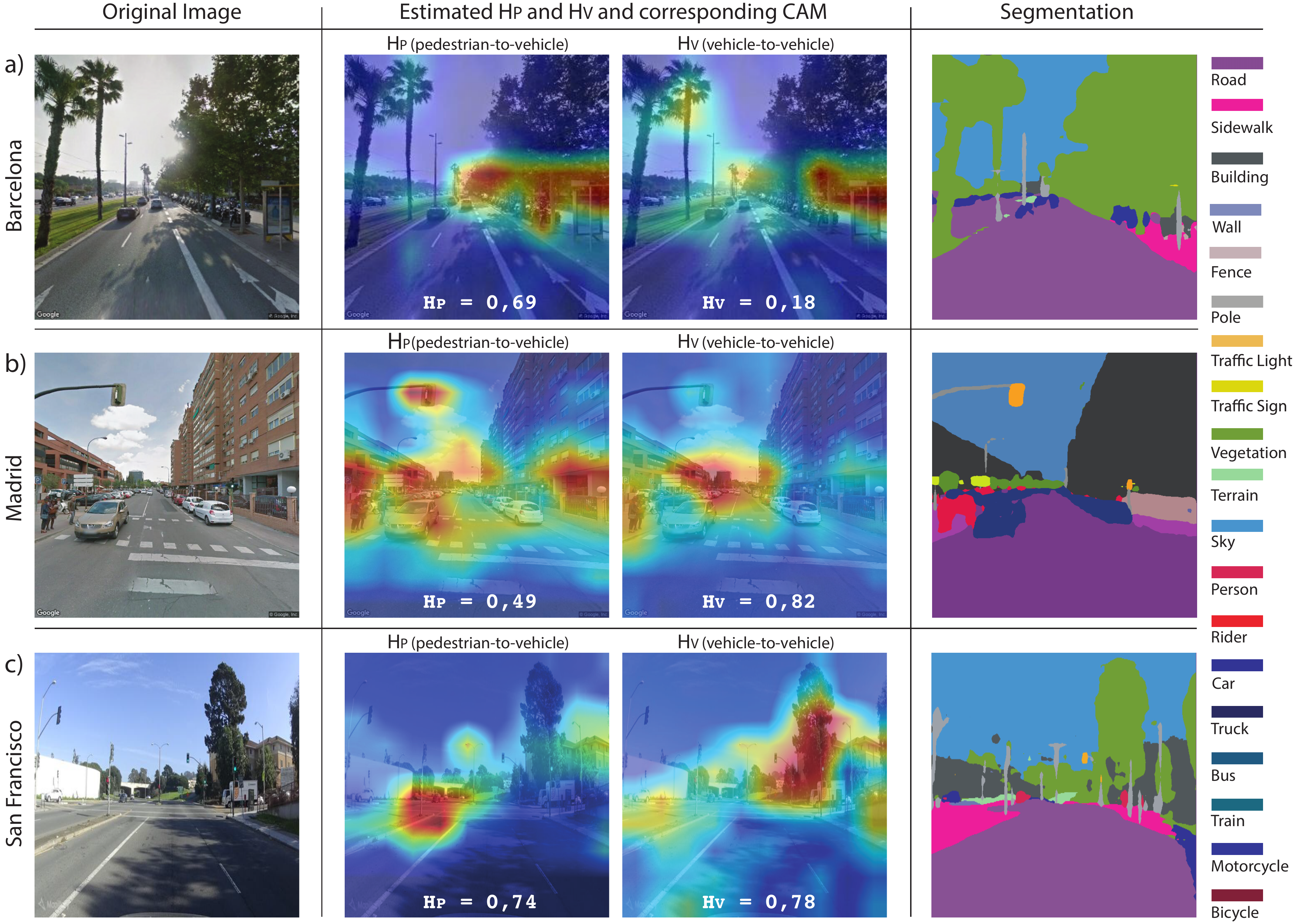}
	\end{center}
	\caption{{\bf Deep Learning approach: classification, segmentation and interpretability.} The figures display image examples from Barcelona, San Francisco and Madrid, one location per row.  First column shows the original street view image. Second and third columns correspond to the obtained CAM for pedestrian and vehicle datasets, respectively. The last column corresponds to the outcome of the segmentation task. The example in Barcelona location (top row) is classified as dangerous for pedestrians (note the score in each picture), but safe for vehicles. The second example, corresponding to a Madrid location, is classified as dangerous for vehicles, but safe for pedestrians. Finally, the third example, corresponds to a San Francisco location. Notice that, in this last case, the location is dangerous for both pedestrian and vehicle, but the CAM highlights different regions: areas increasing the hazard for pedestrians may not coincide with those increasing hazard for vehicles. Images courtesy of Google, Inc. and Mapillary.}
	\label{fig:CamAndSegmentation}
\end{figure*}

\cb{
\subsection{Hazard index Estimation} 
We begin the results section by assessing how well our trained ResNet performs the required classification task for the six datasets we have defined, considering the cities of Barcelona, Madrid, and San Francisco. Images belonging to the `dangerous' class are defined as positive, while those belonging to the `safe' class are defined as negative. In the training stage, the parameter $c$ of the loss function was experimentally assigned as 1. For our results, we focus on the following measures: recall, precision and accuracy; and the indicators: FP (False positives), TP (True Positives), TN (True Negatives) and FN (False negatives). Recall refers to the fraction of samples detected as dangerous over the total number of dangerous samples in the dataset (TP over TP+FN). Precision is the fraction of the true dangerous points detected, over the number of points detected as dangerous by the ResNet (TP over TP+FP). Accuracy measures how good the system is at detecting dangerous points (TP+TN over all the samples).

%\begin{equation}
%    Accuracy=\frac{TP+TN}{TP+TN+FP+FN}
%    \label{acc_eq}
%\end{equation}
%
%\begin{equation}
%    Precision=\frac{TP}{TP+FP}
%    \label{pr_eq}
%\end{equation}
%
%\begin{equation}
%    Recall=\frac{TP}{TP+FN}
%    \label{recall_eq}
%\end{equation}

As we can see in Table~\ref{tab:accuracy}, the obtained accuracy is outstanding for all datasets, considering that the CNN training stage relies only on visual information, along with a binary tag indicating the occurrence (or not) of an accident within a 50m radius (sensitivity with respect to radii is discussed in Section~S4.1 and Figure~S7 of the SI). As illustrated examples of hazard index estimation, see the scores in the central columns of Figure~\ref{fig:CamAndSegmentation}. }

\begin{table*}[th]
  \centering
\begin{tabular}{l|c|c|c|c|c|c|c}
& Recall & Prec. & Acc. & FP & TP & TN & FN \\
\hline
\hline
Barcelona $P$ & 0.86 & 0.72 & 0.75 & 17.8\% & 45.4\% & 29.8\% & 7\%\\
Barcelona $V$ & 0.77 & 0.84 & 0.82 & 7.1\% & 37.9\% & 44.1\% & 10.9\%\\
\hline
\hline
Madrid $P$ & 0.76 & 0.75 & 0.75 & 12.4\% & 37.5\% & 38\% & 12.1\%\\
Madrid $V$ & 0.73 & 0.74 & 0.75 & 12\% & 35.2\% & 40.1\% & 12.7\%\\
\hline
\hline
San Francisco $P$ & 0.63 & 0.81 & 0.76 & 6.6\% & 29\% & 47.7\% & 16.7\%\\
San Francisco $V$ & 0.61 & 0.82 & 0.74 & 6.3\% & 30.1\% & 44.7\% & 18.9\%\\
\end{tabular}
  \caption{Results of the Deep Learning approach for accident prediction, considering a 50 meters radius. Rows labelled as $P$ and $V$ correspond to pedestrian-to-vehicle and vehicle-to-vehicle accident dataset, respectively. Results for other radii can be seen on Table~S1 of the SI.}
  \label{tab:accuracy}
\end{table*}

\cb{Additionally, we compared the performance of different ResNet and other state-of-the-art architectures against the Barcelona dataset. Metrics like F1-score, area under the Precision and Recall (PR) curve, and the area under the Receiver Operating Characteristic (ROC) curve were used for comparison as well. The F1-measure provides a balance between precision and recall in a single score:
\begin{equation}
    F1=2\cdot\frac{precision\cdot recall}{precision+recall}
\end{equation}
Whereas the PR curve represents the balance between the measures precision and recall through different thresholds between 0 and 1. The ROC curve plots the false positive rate versus the true positive rate through different thresholds, like the PR curve. The results presented in Table~\ref{tab:dl_models} show that the ResNet-v2-50 offers the highest performance for this particular image classification task. }

\cb{Discerning between safe and dangerous locations in a binary fashion might be limiting in several practical scenarios, such as the prioritization of urban interventions to improve pedestrian safety. To assess to what extent we can produce finer results, we have also implemented the method in \cite{frank2001simple} to learn an ordinal regressor. In this case, the Barcelona pedestrian dataset was divided in four rating classes: \textsl{no-danger}, \textsl{mild-danger}, \textsl{danger} and \textsl{high-danger}. Images tagged as `no-danger', correspond those images where no accidents were observed. Images in the class `mild-danger' had one accident nearby, images in class `danger' have between 2 and 5 accidents nearby. Finally, images belonging to class `high-danger' have more than 5 accidents in their vicinity. The dataset proportions were approximately 85k, 34k, 40k and 17k images samples, respectively. The method in \cite{frank2001simple} relies on several binary classifiers. We used our same ResNet architecture for each of those binary classifiers. After training, we obtained a balanced accuracy of 0.47 (with a the dummy classifier accuracy of 0.25) which is comparable to the performance reported in \cite{song2018farsa} for a similar task. That is, the ResNet architecture can also provide competitive results for a finer assessment of pedestrian safety.} 

\begin{table*}[t]
  \centering
\begin{tabular}{l|c|c|c|c|c|c}
Model & Acc.  & Prec. & Rec.l & F1-Score & PR & ROC  \\
\hline
\hline
VGG16 \cite{simonyan2014very}& 0.61 & 0.58 & 0.96 & 0.72 & 0.78  & 0.59 \\
\hline
VGG19 \cite{simonyan2014very}& 0.68 & 0.73 & 0.62 & 0.67 & 0.77  &  0.68 \\
\hline
Inception-V3 \cite{szegedy2016rethinking} & 0.70 & 0.70 & 0.75 & 0.72  &0.79  &  0.70 \\
\hline
Inception-V4  \cite{szegedy2017inception}& 0.57 & 0.80  &  0.24 & 0.37 &  0.72 &  0.59 \\
\hline
Mobilenet \cite{howard2017mobilenets} & 0.62 & 0.77  &  0.39 &  0.52  &0.74 & 0.63 \\
\hline
ResNet-v1-50 \cite{resnet} & 0.61 & 0.80 & 0.35 &  0.49 &0.75 & 0.63  \\
\hline
ResNet-v1-101 \cite{resnet}& 0.59 & 0.56  & 0.99 &  0.71 &0.78  & 0.57  \\
\hline
ResNet-v1-152 \cite{resnet}& 0.67 & 0.71 & 0.62 & 0.66  & 0.76 & 0.67  \\
\hline
ResNet-v2-50 \cite{resnet_v2}& \textbf{0.75} & 0.72  & 0.87 &  \textbf{0.78} & \textbf{0.82}  & \textbf{0.74}  \\
\hline
ResNet-v2-101 \cite{resnet_v2}& 0.72 & 0.75 & 0.70 & 0.72  & 0.80 & 0.72  \\
\hline
ResNet-v2-152 \cite{resnet_v2}& 0.72 & 0.74 & 0.72 & 0.73 & 0.80  &  0.72 \\
\end{tabular}
  \caption{\cb{Results of the Deep Learning approach for accident prediction, considering different classification architectures.}}
  \label{tab:dl_models}
\end{table*}

%\subsection{Accidents dataset description} 

%%%%%%%%%%%%%%%%%%%%%%%%%%%%%%%%%%%%%%%%%%%%%%%%%%%%%%%%%
%%%%%%%%%%%%%%%%%%%%%%%%%%%%%%%%%%%%%%%%%%%%%%%%%%%%%%%%%
%\section{Results}
%\label{sec:results}

\subsection{Urban hazard landscape} 
The first remarkable outcome of the described methodology (in particular, Section~2.2) is a fine-grained map of hazard indices throughout the cities under study. The Deep Learning approach, together with the short distance intervals between consecutive images, allows us to quantify the safety of all city locations at a microscopic level, i.e. every 15 meters approximately (see Figures~S3 and S4 in the SI), independently of whether accidents have occurred at a given site or not. %Assuming that accidents increase proportionally with the flow of pedestrians and vehicles at a location, and thus their chances to coincide in a shared space, our method provides crucial information about the hazard potential that a location has if the flow of pedestrians and vehicles were incremented (e.g. due to the construction of a shopping centre, hospital or school).

\begin{figure*}
  \begin{center}
 	 \includegraphics[width=1\columnwidth]{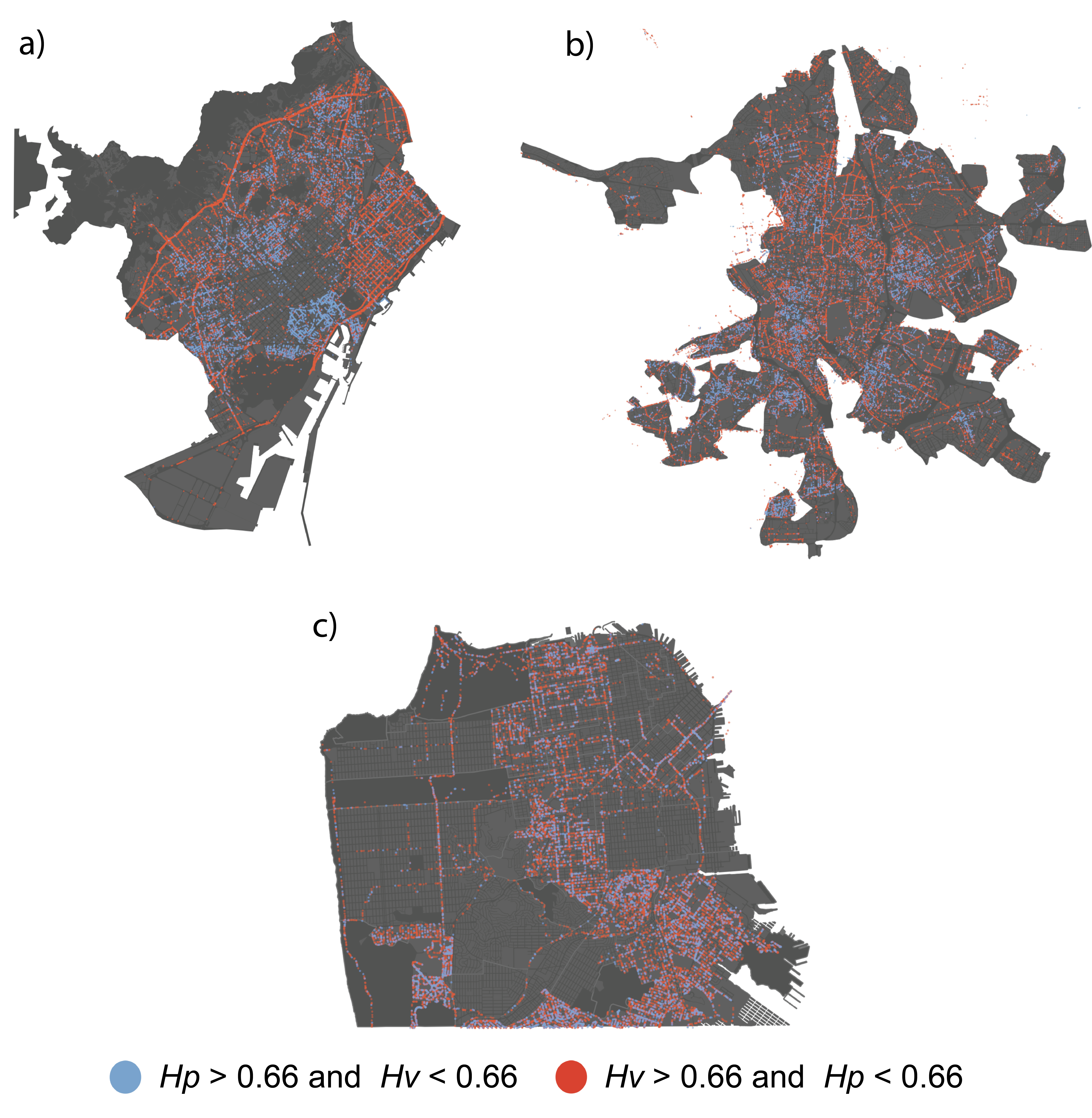}
	\end{center}
	\caption{{\bf Spatial distribution of hazard index}. Distribution of high-hazard points for pedestrians and vehicles across all three cities of study. Points displayed are those for which hazard is high for pedestrians (vehicles) but not for vehicles (pedestrians).}
	\label{fig:hazardHistograms}
\end{figure*}

To give a complete picture of hazard for pedestrians and vehicles, and to highlight their differences, Figure~\ref{fig:hazardHistograms} shows the spatial distribution of points that were identified as very hazardous for pedestrians ($H_P \geq 0.66$), but with low-to-moderate hazard for vehicles ($H_V < 0.66$), and vice-versa. As can be seen, in both Madrid and Barcelona, areas of high hazard for pedestrians alone are highly concentrated in the denser, older city centers. High levels of vehicle hazard tend to be distributed around arterial roads, as well as some distinct neighborhoods (e.g. Sant Mart\'i-Poble Nou, middle right corner in Barcelona). San Francisco presents an interesting case in which the two spatial distributions are nearly homogeneous. This can likely be explained by the bias towards residential, medium-density areas in our image coverage for the city (see Materials and Methods for further discussion). Notably, we lacked image coverage in high-density downtown San Francisco, as well as peripheral low-density districts. With the inclusion of such zones, it is possible that clearer spatial patterns would emerge, although they might be distinct from those of denser European cities like Barcelona and Madrid \cite{louf2014typology}. Nevertheless, it should be noted that competitive levels of precision and accuracy were still achieved in San Francisco, indicating that our method is robust to relatively homogeneous training data. \cb{Furthermore, it shows that the classifier need not only be applied to comprehensive collections of images from an entire city, but can function well on sufficiently rich, spatially homogeneous samples of images}. Separate visualizations for pedestrian and vehicle hazards are available in the SI, Figure~S3.

Worth highlighting, there has been no previous attempt to associate a given street image with traffic hazard levels --unlike other urban attributes (e.g. beauty \cite{quercia2014shortest,naik2017pnas}, or security \cite{naik2014streetscore}). Here, we do so under the assumption that street-level imagery is a good proxy for both the structural and perceptual complexity of the city landscape. Typically, traffic-related risk is either aggregated to the macro-level (neighborhoods, census tracts, even counties)\cite{huang2010county,ukkusuri2012role,chen2016effects}, or painstakingly micro-tailored to very specific settings (e.g. considering only zebra-crossings \cite{olszewski2016pedestrian}). However, initiatives like Vision Zero, involving governments and organizations worldwide, demand new streams of data and methodologies that help address the street safety challenge at the finest level {\it and} at scale. This is achieved here combining images and accident data.

\subsection{Mapping safety to scene composition}
The second (segmentation) and third (Class Activation Mapping, CAM) processing steps complete the data analysis pipeline, linking hazard indices, $H_{P}$ and $H_{V}$, to specific objects found in street-level images. In practice, such link is established combining the information in the central and right columns of Figure~\ref{fig:CamAndSegmentation}. Mapping each pixel label (e.g. ``road'', ``sidewalk'', etc.) to its corresponding activation level (heatmap in central columns of Figure~\ref{fig:CamAndSegmentation}) provides a quantification of the contribution of that pixel to the overall hazard score of the image. Thus, at the city level, we can obtain a global perspective of the categories that most contribute to the hazard index. 

\begin{figure}[t]
  \begin{center}
 		 \includegraphics[width=1\columnwidth]{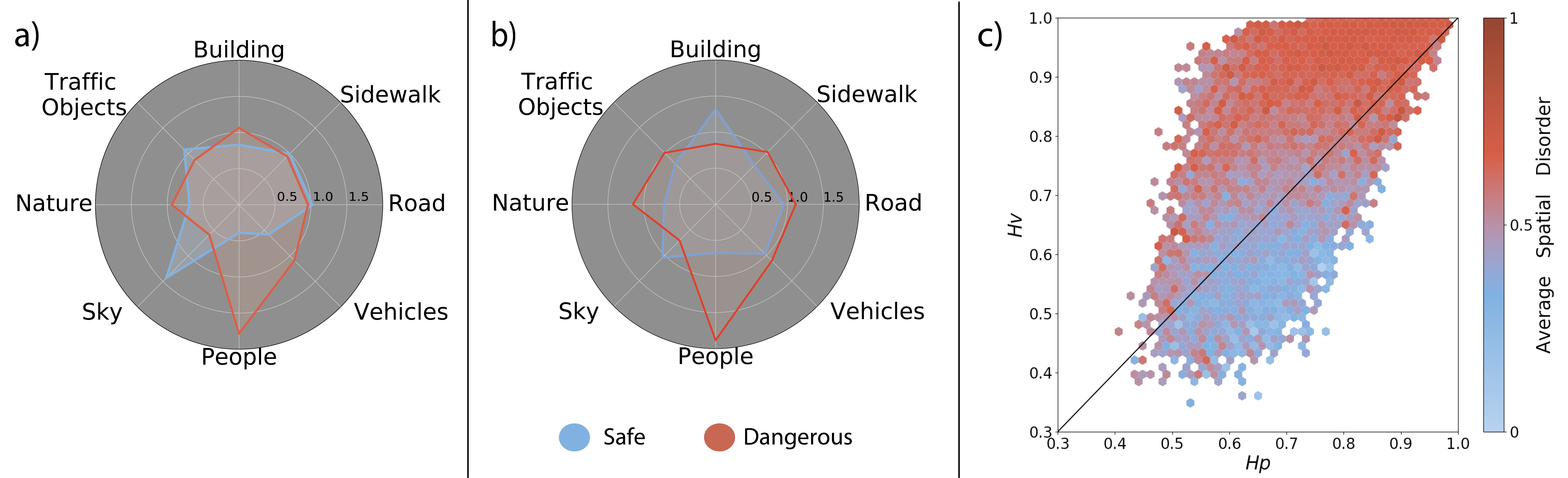}
	\end{center}
	\caption{{\bf Hazard level interpretability.} {\bf Top:} Radar plots showing the level of object fixation of the CAM model for pedestrian (a) and cars (b). For both, the blue area corresponds to images classified as safe ($H < 0.33$), while scenes classified as dangerous ($H > 0.66$) are mapped on the plot as red. To build these radars, each individual image is mapped to the radar categories (a relevant subset of those detected by the segmentation task), and the average of such mappings is shown. {\bf (c)} The plot shows the triple relationship between $H_{P}$, $H_{V}$ and the color-coded level of disorder (adapted from \cite{haralick1973textural}) --which increases towards warmer colors as the levels of hazard increase. The plot corresponds to Barcelona.}
	\label{fig:radar}
\end{figure}

\cb{
Figure~\ref{fig:radar} (panels a and b) illustrates this for the central area of Barcelona. These radar plots show the level of object fixation of the CAM model for pedestrians (a) and cars (b). In both cases, the blue line represents safe scenes ($H < 0.33$), while dangerous ones ($H > 0.66$) are shown in red. Specifically, we plot the ratio between the amount of CAM fixation on a given category (in safe and dangerous scenes), with respect to the CAM fixation on that category across all the images of the dataset. Thus, values below 1 in the radar plots are underrepresented, while those above 1 are overrepresented. We would like to highlight that we have restricted the analysis to the city center, to avoid an exaggeration of the presence of natural elements (vegetation and sky) in low accident risk images. 
Remarkably, the presence of people in a scene is correlated to a dangerous classification for both vehicle-to-pedestrian and vehicle-to-vehicle predictions. Low buildings and/or wide streets (tantamount to a clear vision of the sky) correlate to safer scenes for pedestrians, whereas the presence of buildings implies a safer environment for vehicles. Also, the absence of vegetation, such as trees, could be contributing to a safe classification for vehicles.
}

Radar plots for Madrid (see SI, Fig.~S5) show high resemblance to the Barcelona ones, while those for San Francisco (Fig.~S6) show completely different patterns: for pedestrians, the presence of sidewalks --and not people-- is identified as the strongest driver for high $H_{P}$. Again, the distinct layouts and walking habits of European and North American cities may be directly related to these emergent patterns. 

Moving further, we can relate hazard levels to the scene complexity. While the radar plots show interesting information, they are blind to specific scene compositions in urban scenes, i.e. whether categories appear in a clustered or fragmented way. To grasp this information, we quantify scene disorder ($SD$) as defined in Equation~\ref{spadisorder}, see Methods above. Figure~\ref{fig:radar}c shows an hexbin scatter plot of hazard indices ($H_{V}$ against $H_{P}$), with a color-coded third dimension that corresponds to scene disorder, normalized in the range $[0, 1]$. A first observation is that $H_{P}$ and $H_{V}$ are positively correlated. More interestingly, it is clear that more complex scenes (warmer colors) correspond to more dangerous ones. In Figure~S5c of the SI, an even clearer trend is shown for Madrid. On the other hand, the level of disorder in San Francisco scenes is high when $H_{P} \approx H_{V} \approx 1$, but not clearly related to either $H_{P}$ or $H_{V}$ for the rest of values, see Figure~S6c. All in all, the connection between image complexity and hazard (especially for vehicles) suggests that more research is needed in this direction. While certain distractions are very explicit (e.g. attending the mobile phone), the perils of scene disorder are subtle and implicit (in the sense that they are not obvious on visual inspection).

\subsection{An informed guide to pedestrian safety improvements}
A precipitate analysis of Figure~\ref{fig:radar} may render unfeasible interventions: substitution of built space with larger green areas, building height reduction, or street widening would suffice to improve pedestrian safety, but they do not represent a realistic approach. Instead, we resort on the greedy strategy developed in Section~\ref{sec:greedy} to propose interventions conducive to scene alterations that diminish $H_P$ and $H_{V}$ most.

\begin{figure*}[h!]
  \begin{center}
		\includegraphics[width=0.9\columnwidth]{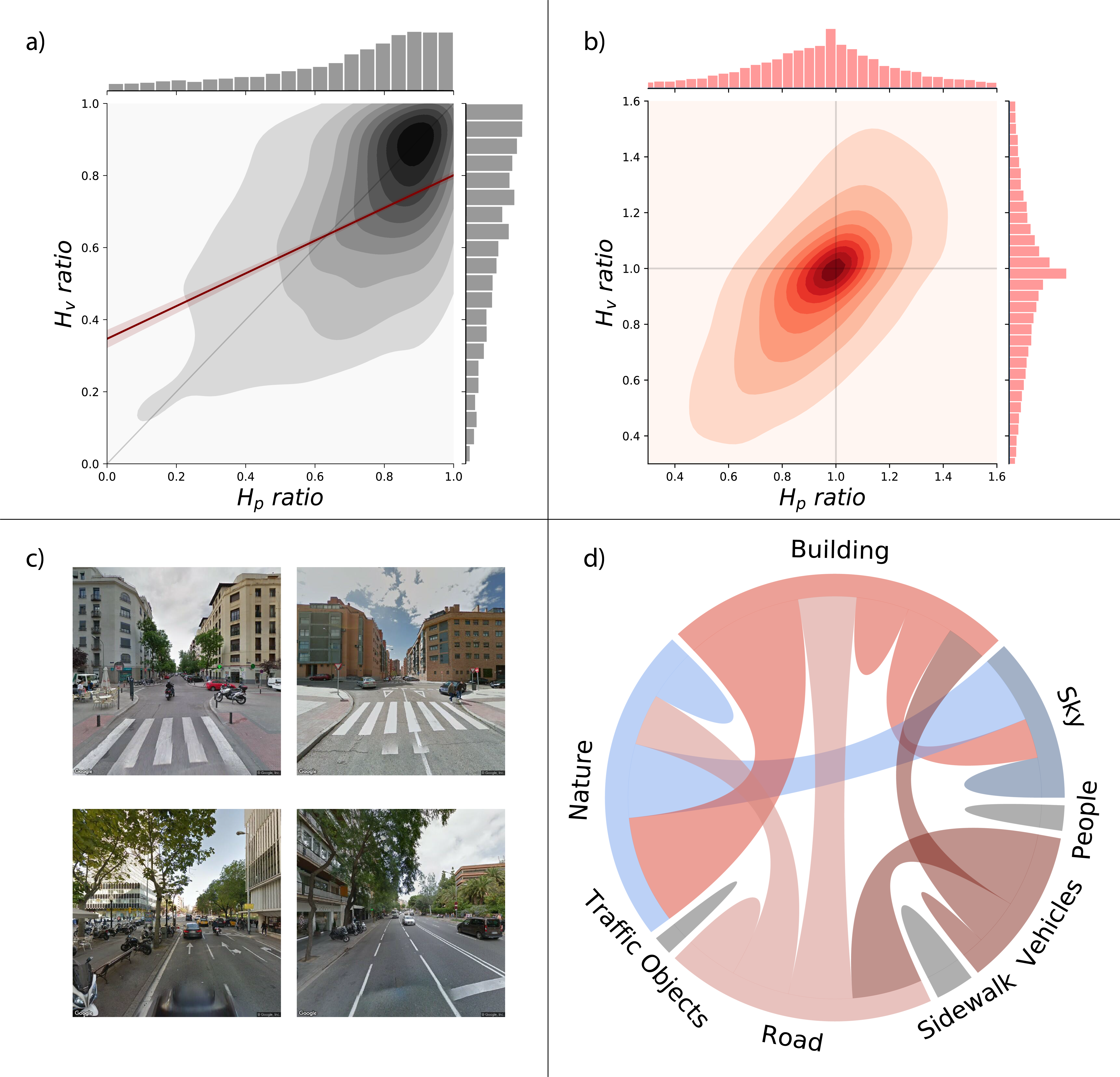}
	\end{center}
	\caption{{\bf Hazard reduction: results. (a)} Expected improvement for pedestrian and vehicle hazards, with respect to their original values. The horizontal axis corresponds to the ratio between the improved and the original pedestrian hazard index, $\tilde{H}_P / H_{P}$; while the vertical axis represents the equivalent ratio for vehicles, $\tilde{H}_V / H_{V}$. Grey intensity represents the density of observations in a given area of the plot. \cb{{\bf (b)} Expected improvement of a dummy $k$-nn algorithm that only considers similarity between images. This can be regarded as a baseline for results in panel (a)} {\bf (c)} Examples of original and mirror images in Barcelona and Madrid. {\bf (d)} Chord diagram representing an aggregate overview of proposed interventions in Barcelona. The most notable outcome from the diagram is the propensity to reduce the space allotted to roads and buildings, exchanging it emptier, greener scenes.}
	\label{fig:heuristics}
\end{figure*}

Figure~\ref{fig:heuristics}a shows the results of the application of this optimization to the set of images in Barcelona (Figure~S8 in SI for Madrid and San Francisco). In some occasions the hazard index cannot be reduced (points near the $(1,1)$ coordinate). And yet, many locations present a potential to decrease the hazard levels, even observing, for some scenarios, extreme improvements (points near the $(0,0)$ coordinate). The grey intensity in Fig.~\ref{fig:heuristics}a reflects the density of observations in that area. \cb{To provide a baseline for comparison, panel b shows alternative results considering a dummy $k$-nn regressor, that does not take our hazard index into account.  Ratios larger than 1 indicate an increase in $H_{V}$ or $H_{P}$, and ratios lower than 1 indicate a decrease. The average in both dimensions is close to zero, evidencing that, with a dummy regressor, we have no guarantee of reducing either pedestrian or vehicle hazard.} Figure~\ref{fig:heuristics}c shows a selection of two targets and their most similar mirror image, illustrating some common interventions proposed by the heuristic (more examples, for the three cities under study, can be found in Figure~S9 of the SI). Visually, all of them seem to point at simplifications of the original image -- mostly removing objects on sidewalks.

Finally, Figure~\ref{fig:heuristics}d provides a visual overview of the most frequent interventions predicted by our optimization scheme, in the case of Barcelona. The color of the link connecting two categories expresses the source of that link. The most notable changes point --perhaps unsurprisingly-- to the need to reconfigure urban scenes towards greener and wider spaces: indeed, both categories 'road' and 'building' contribute largely to 'nature', while the latter does the same towards 'sky'. Madrid presents an almost identical trend, while San Francisco shows a less clear pattern (although the relevance of 'nature' and 'sky' is still clear). Both diagrams are available in the SI, Figure~S10. Overall, the estimations and insights from the panels in Fig.~\ref{fig:heuristics} can provide initial indications to urban planners about achieving potential reductions of a local hazard score, both in terms of which items could be removed or relocated.

%%%%%%%%%%%%%%%%%%%%%%%%%%%%%%%%%%%%%%%%%%%%%%%%%%%%%%%%%
%%%%%%%%%%%%%%%%%%%%%%%%%%%%%%%%%%%%%%%%%%%%%%%%%%%%%%%%%
\section{Discussion}
\label{sec:discussion}
%This should explore the significance of the results of the work, not repeat them. A combined Results and Discussion section is often appropriate. Avoid extensive citations and discussion of published literature.
As cities become increasingly populated, the interactions among pedestrians and motorized vehicles become permanent. This translates into a growing number of pedestrian-vehicle accidents. Complementary to the efforts by urban planners, public authorities and sensor technology designers, we present here an automated scheme that exploits a wide range of Computer Vision methods (classification, segmentation and interpretability techniques) to reduce traffic-related fatalities. The proposed processing pipeline, conveniently fed with rich sources of open data, renders an holistic characterization of a city's hazard landscape, capturing the physical (scene structure) and perceptual (scene complexity) characteristics from a car driver's point of view. Beyond its informative value, the hazard landscape provides actionable insights to planners.

The main strength of our proposal lies in its simplicity, and its potentially universal applicability out of a comprehensive street image collection and a rich accident dataset. Even crowd-sourced imagery, which is unavoidably diverse and often sparse, provides a solid starting point to quantify safety at a below-segment level. A global, automated, data-driven endeavour towards improving pedestrian safety is not out of reach, considering the advances in cities' public data portals, and the wide coverage of proprietary services like Google Street View or open initiatives like Mapillary.

Our approach opens a promising line of development. \cb{The hazard landscape is defined at an unprecedented, sub-segment resolution level --roughly a hazard score every 15 meters-- through an automated and scalable classification process}. This is well beyond macroscale approaches (e.g. crash hotspots), and extends the emphasis on intersections \cite{hu2018dangerous}. Such fine-grained map adds a valuable geoinformation layer to those already in use --traffic and pollution levels \cite{xu2019unraveling}, land and underground transportation systems, crime, etc.-- enabling better route design: safe paths, along with clean, beautiful, or shortest ones.

Additionally, segmentation and interpretability methods unveil the relationship between potential danger and specific objects in urban scenes. What's more, the disposition of those objects is related to hazard indices, adding a perceptual-attentional link to other possible concomitant variables that affect vehicle and pedestrian safety. Along this line, our work \cb{can be used in conjunction with other similar pipelines, such as \cite{song2018farsa}, which automates road safety assessment in terms of infrastructure and estimates road attributes, or may contribute to more focused analysis, relating what a person pays attention to while driving \cite{palazzi2018predicting}. Additionally, further information such as temporal accident data, or factors known to influence accident rate (e.g. weather, lighting condition, distraction, asphalt conditions, road signaling) could be included by using, for instance, a multi-branch convolutional neural network, to obtain a richer prediction model.}

On the other hand, the step from descriptive (hazard landscape) to actionable insights paves the way to automatized, computer-aided prioritization of urban interventions. The proposed heuristic towards safety improvements can serve as a novel tool for planners and policy makers, and might trigger the development of more sophisticated approaches such as the use of Generative Adversarial Networks to produce virtual, plausible alternatives to target scenes (seeking for instance ``safe-fication'', instead of ``beautification'' \cite{kauer2018mapping}). These techniques could be complemented with intervention cost quantification, considering as well cost-safety gain trade-offs.

%CITAS SUPPLEMENTARY:
%\cite{bcn2019acc,mad2019acc} 
%\cite{tims2019} 
%\cite{OpenStreetMap} 
%\cite{boeing2017osmnx}
%\cite{frank2001simple}

%%%%%%%%%%%%%%%%%%%%%%%%%%%%%%%%%%%%%%%%%%%%%%%%%%%%%%%%%
%%%%%%%%%%%%%%%%%%%%%%%%%%%%%%%%%%%%%%%%%%%%%%%%%%%%%%%%%

%\section{Conclusions}
%\label{sec:conclusions}
%The main conclusions of the study may be presented in a short Conclusions section, which may stand alone or form a subsection of a Discussion or Results and Discussion section.

%%%%%%%%%%%%%%%%%%%%%%%%%%%%%%%%%%%%%%%%%%%%%%%%%%%%%%%%%
%%%%%%%%%%%%%%%%%%%%%%%%%%%%%%%%%%%%%%%%%%%%%%%%%%%%%%%%%

%\section{Glossary}
%Please supply, as a separate list, the definitions of field-specific terms used in your article.

%% The Appendices part is started with the command \appendix;
%% appendix sections are then done as normal sections
%\appendix

\section*{Acknowledgements}
All authors acknowledge financial support from the Direcci\'on General de Tr\'afico (Spain), Project No. SPIP2017-02263, as well as TIN2015-66951-C2-2-R and RTI2018-095232-B- C22 grants from the Spanish Ministry of Science, Innovation and Universities (FEDER funds). CB and DR acknowledge as well the support of a doctoral grant from the Universitat Oberta de Catalunya (UOC). CB, DM and AL acknowledge the NVIDIA Hardware grant program. Street network data copyrighted OpenStreetMap contributors and available from https://www.openstreetmap.org.

%% \section{}
%% \label{}

%% References
%%
%% Following citation commands can be used in the body text:
%% Usage of \cite is as follows:
%%   \cite{key}          ==>>  [#]
%%   \cite[chap. 2]{key} ==>>  [#, chap. 2]
%%   \citet{key}         ==>>  Author [#]

%% References with bibTeX database:

%\bibliographystyle{model1-num-names}
%\bibliography{sample}

%% Authors are advised to submit their bibtex database files. They are
%% requested to list a bibtex style file in the manuscript if they do
%% not want to use model1-num-names.bst.

\end{document}